# Fully Convolutional Deep Network Architectures for Automatic Short Glass Fiber Semantic Segmentation from CT scans

Tomasz Konopczyński[1, 2, 3], Danish Rathore[2, 3], Jitendra Rathore[5], Thorben Kröger[1],
Lei Zheng[2], Christoph S. Garbe[3], Simone Carmignato[5], Jürgen Hesser[2, 3, 4]

[1]Volume Graphics GmbH, Heidelberg, Germany, e-mail: tomasz.konopczynski@volumegraphics.de
[2]Department of Radiation Oncology, University Medical Center Mannheim, Heidelberg University, Germany
[3]Interdisciplinary Center for Scientific Computing (IWR), Heidelberg University, Germany
[4]Central Institute for Computer Engineering (ZITI), Heidelberg University, Germany
[5]Department of Management and Engineering, University of Padova, Vicenza, Italy

**Abstract**
We present the first attempt to perform short glass fiber semantic segmentation from X-ray computed tomography volumetric datasets at medium (3.9 µm isotropic) and low (8.3 µm isotropic) resolution using deep learning architectures. We performed experiments on both synthetic and real CT scans and evaluated deep fully convolutional architectures with both 2D and 3D kernels. Our artificial neural networks outperform existing methods at both medium and low resolution scans.

## 1 Introduction

Reliable information about fiber characteristics in short-fiber reinforced polymers (SFRP) is much needed for the process of optimization during the product development phase. The main characteristics of interest are fiber orientation, fiber length distribution and the percent composition in the product. The influence of these characteristics on the mechanical properties of SFRP composites is of particular interest and significance for manufacturers. The recent developments of X-ray computed tomography (CT) for nondestructive quality control enabled the possibility to scan the materials and retrieve the 3D spatial information of SFRPs.

In recent years, deep learning methods have revolutionized various fields to which they have been applied. In computer vision, fully convolutional networks (FCN) have become the architecture of choice for the task of semantic segmentation [1]. In particular, FCNs have also been successfully applied to a number of medical 3D X-ray datasets [2]. This motivates us to apply these ideas to the task of fiber semantic segmentation of SFRP. Semantic segmentation is the task of assigning each voxel a label out of a predefined set of categories. The problem of fiber semantic segmentation is therefore a binary classification with 'fiber' and 'non-fiber' categories. This information can be used for further analysis like single fiber segmentation or directly to compute the fiber volume ratio and consequently the fiber weight in a specimen. However, the spatial resolution of a scan is a limiting factor and makes it difficult for the standard segmentation methods to work properly. The currently used methods have difficulties with fibers that are approx. 2 voxels in diameter or less. The glass fibers we use are approx. 13 µm in diameter. Therefore, we consider scans at a medium resolution (MR) of 3.9 µm, which is a limit for which the standard methods work reasonably well, and a low resolution (LR) of 8.3 µm for which standard methods usually fail.

We implemented and evaluated a residual [3] version of a FCN with 2D and 3D kernels. Both networks are capable of segmenting fibers better than our baseline algorithms. We also find that training on synthetic volumetric data, but predicting on real scans gives good results. This means that one does not need a large amount of annotated data to train the networks, which is often an argument against using deep learning. We compare our results with Otsu thresholding [4], Hessian based Frangi vesselness measure [5], and our internal implementation of a classical machine learning setup with a random forest [6] using a set of predefined features.

## 2 Related Work

One of the easiest and most widely used algorithms for semantic segmentation is Otsu thresholding [4], which is a gray value based histogram method. The reliability of global histogram-based methods is limited due to noise and brightness variations over the image. To overcome this problem, slice-wise circular Hough Transform or Circular Voting [7, 8] can be used. These methods take advantage of the geometrical information about the scanned part, e.g. the radius of a tubular structure, in order to search for circular or elliptical structures in 2D slices. However, these algorithms do not scale well to 3D data. The most common methods to segment tubular-looking structures in 3D data use Hessian based filters. One of the first Hessian based methods, Frangi vesselness filter [5], was initially developed for segmenting vessels in biomedical images and is now commonly used as a preprocessing step on 3D CT data. For fiber-reinforced polymers, a priori information such as the radius of fibers or expected orientation distribution can be incorporated into the algorithm. The method developed by Zauner et al. [9] is a good example of using a priori knowledge dedicated to the analysis of fiber-reinforced polymers.

## 3 Network Architecture

For this work, as mentioned in the introduction, we decided to use the framework of fully convolutional architecture with residual units. A residual unit consists of two convolutional layers followed by a batch normalization and a non-linearity layer. We have not used pooling layers in our architectures, not to decrease the resolution of already very small fibers. For the non-linarites we decided to use the rectifier linear unit (ReLU). We evaluated a slice-wise 2D and a 3D model. We also compared a relatively shallow deep model and a deeper version of it (with more convolutional layers). We have termed these models as "Shallow model" and "Deep model". The example diagram representation of these models is shown in figure 1.

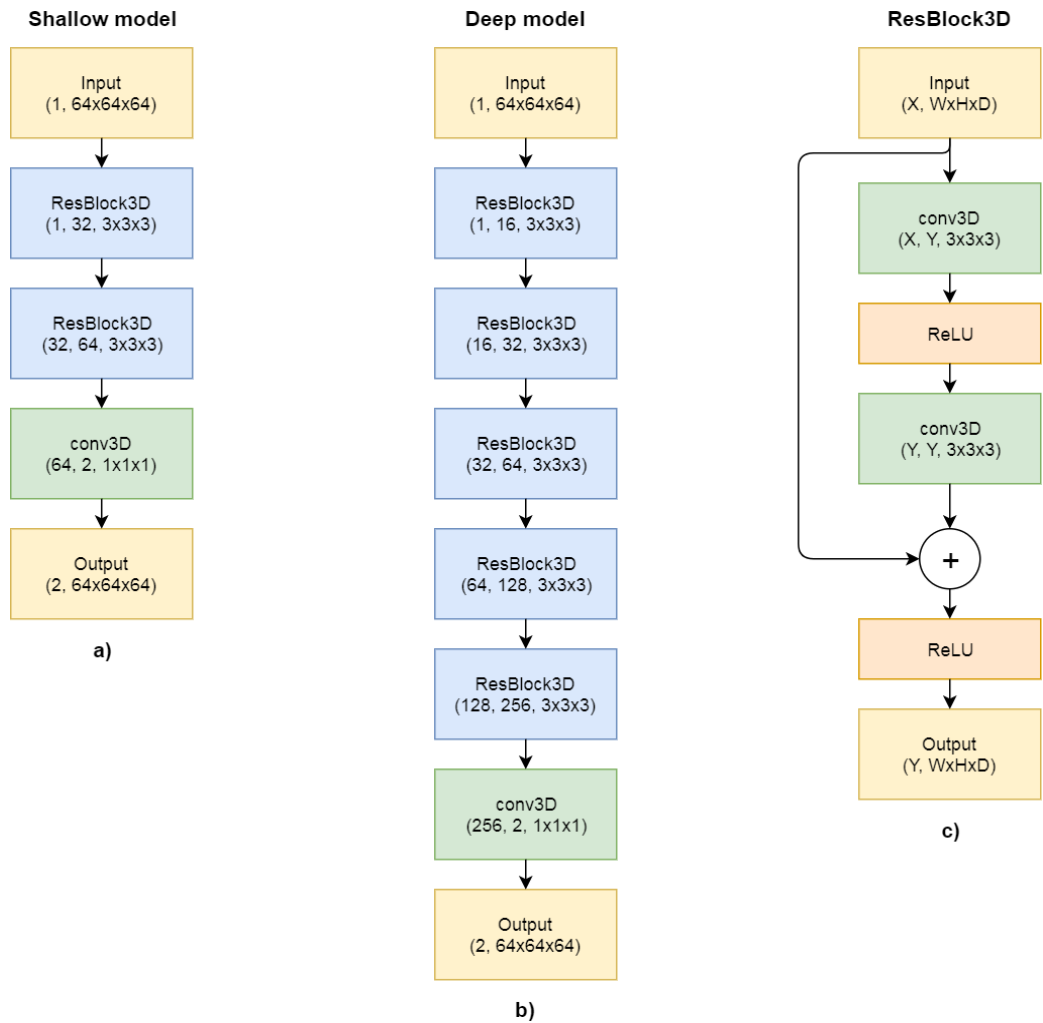

**Figure 1:** Diagram representation of the models used for learning from 3D MR patches. (a) Shallow model and (b) Deep model version of the architecture. In both cases the input is a one-channel block of dimension $64 \times 64 \times 64$. Going through the network, the number of channels increases creating a feature maps of the patch. It is done by convolving the input with a number of $3 \times 3 \times 3$ kernels (weights) inside the residual blocks (ResBlock3D). The final convolutional layer maps the last feature maps into a two-channel output. One for the fibers, and another for the background (epoxy). c) The architecture of a residual block. It consist of two convolutional layers with a non-linearity in between and an identity connection. $X$ is the number of channels of the input, and $Y$ is the desired number of channels on the output. $W$, $H$ and $D$ are correspondingly width, height and depth of a patch.

## 4 Dataset

We have used a publicly available dataset of CT scans of SFRP [10]. The parts from which the dataset was created were manufactured by micro injection molding using PBT-10% GF, a commercial polybutylene terephthalate PBT (BASF, Ultradur B4300 G2) reinforced with short glass fibers (10% in weight). The real scans were acquired by a Nikon MCT225 X-ray CT system. The synthetically modeled scans have been generated by our own software. The synthetic scans have the same resolution, same fiber diameter and fiber density as the real scans, but different orientation and length distribution. Finding the length and orientation of fibers is out of the scope of this work. Since the orientation of the fibers in real scans was unknown, we decided to model the synthetic scans with uniformly oriented fibers. We have used two synthetic and two real CT scans, where each volume has corresponding binary ground-truth annotations with voxels annotated as fibers. The real scans are provided only in MR, while the synthetic are in both MR and LR. All the scans have isotropic resolution. The summary of the data is shown in table 1.

**Table 1:** The summary of the data used in this work.

|  | **Resolution** | **Dimensions [voxel]** |
|---|---|---|
| **Real MR 1** | 3.9 μm | $200 \times 260 \times 260$ |
| **Real MR 2** | 3.9 μm | $200 \times 260 \times 260$ |
| **Synthetic MR 1** | 3.9 μm | $627 \times 586 \times 594$ |
| **Synthetic MR 2** | 3.9 μm | $635 \times 603 \times 619$ |
| **Synthetic LR 1** | 8.3 μm | $323 \times 340 \times 349$ |
| **Synthetic LR 2** | 8.3 μm | $323 \times 307 \times 424$ |

## 5  Evaluation metric

We use the Dice score as the evaluation metric for quantitative comparisons [10]. One can use this metric to compare the predicted segmentation on a pixel level with the ground truth (labeled volumes). Defining the binary ground truth labels as a cluster $B$ and the corresponding predicted binary labels as a cluster $B'$, the Dice coefficient index $D$ is defined by

$$D(B, B') = \frac{2 \times |B \cap B'|}{|B| + |B'|} = \frac{2 \times TP}{2 \times TP + FN + FP}$$

where the intersection operation is the voxel-wise minimum operation, and $|\cdot|$ is the integration of the voxel values over the complete image, $TP$ is the true positive, $FP$ false positive and $FN$ false negative. The score varies between 0 and 1, where 1 means a perfect match between the algorithm output and the ground truth mask, and 0 complete mismatch.

## 6  Experiments

First we compared the performance of the deep neural networks with the standard methods, then we examined different setups of the model. We evaluated a 2D versus 3D version of the model and the influence of the number of the convolutional layers (residual blocks). All the models are limited to 8,000 training iterations with a batch size of 3. We used the standard Adam optimizer [11] with a learning rate of 0.001. All the volumes in the dataset have been normalized to have unit variance and zero mean. The patch size of the training data is set arbitrarily, and kept to cover the same region for MR and LR volumes. The patch size is limited by the memory of a GPU card on which the network is trained, that is why we used smaller patches for the 3D models. The LR models were trained on $32 \times 32$ patches in the 2D version and $16 \times 16 \times 16$ volumetric patches in the corresponding 3D version. The MR models were trained on $64 \times 64$ patches in the 2D version and $32 \times 32 \times 32$ volumetric patches in the corresponding 3D version. The patches were randomly flipped and rotated (by 90°, 180° or 270°) during the training phase. While the training was performed on subpatches of a volume, the evaluation was performed and reported on the entire testing volume. The models have been implemented in pytorch [12] and trained on a single GPU Titan X in pascal architecture with 3584 CUDA cores and 12 GB of memory.

### 6.1 Comparison with standard methods

We performed the comparison on one volume (either MR or LR) and evaluated on another one. If the method did not require learning we simply report its performance. We provide the result of finding the best threshold on the particular volume based on the groundtruth, which is the upper boundary for threshold based algorithms. In this experiment we used only the deep models for comparison. An example visual comparison of the methods is shown in figure 2.

For the MR data Otsu thresholding, depending if the data is synthetic or real, is close to the optimal threshold (which is not known a priori). Otsu thresholding is sensitive to the histogram distribution of voxel intensities and in a result, because of slightly different statistics of the synthetic volume, it is not able to find a good threshold for it. The Hessian based method works better on the synthetic data than Otsu and worse on real data. It is so, because fibers in the synthetic data are randomly oriented straight tubes, and therefore have less touching points compared with real data. Random forest, which uses a set of standard features (Gaussian smoothing, Gaussian gradient magnitude, Laplacian of Gaussian, eigenvalues of Hessian of Gaussian and eigenvalues of structure tensor) produces better results than the Hessian based technique and works better for the synthetic data than Otsu. Deep neural networks outperformed the other methods in terms of Dice score at both MR and LR settings when trained on a volume with similar statistics. Especially at LR, deep models achieve a much better result. The deep neural networks perform similarly to Otsu best performance in terms of Dice score for the real MR 1 scan when trained on a synthetic volume. The 3D version of the deep model is working as well as the 2D slice-wise version for MR data, but is significantly superior at LR data.

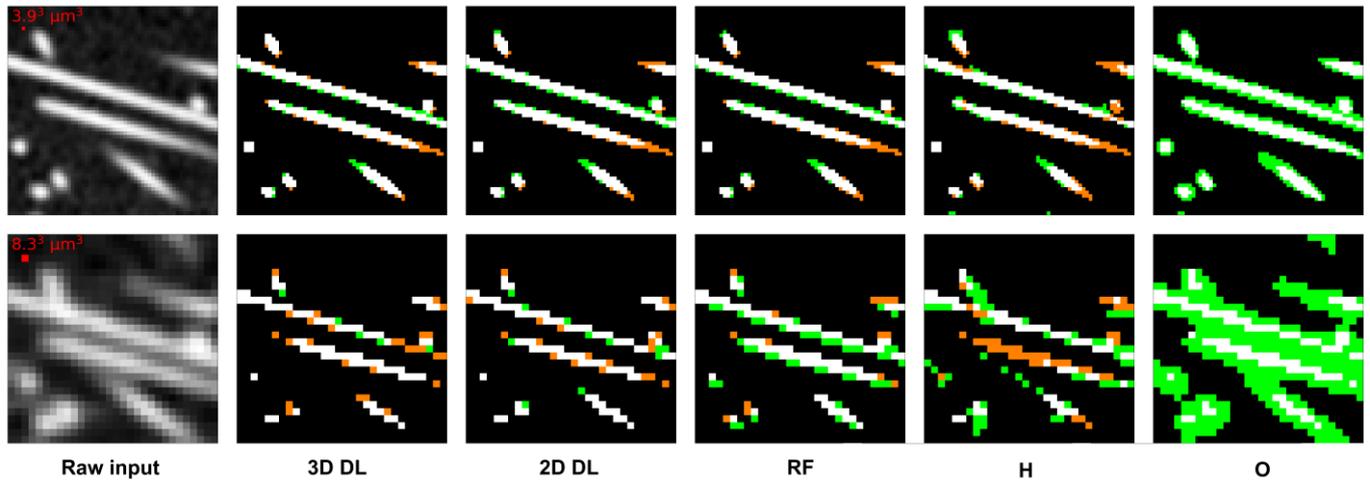

**Figure 2:** Visualization of performance of different methods on a synthetic test scan at MR (first row) and LR (second row). The first column shows a cropped region of a slice of the volume. The following columns correspond to the output of the following algorithms: (3D DL) 3D deep learning network, (2D DL) 2D deep learning network, (RF) random forest, (H) vesselnes measure using Hessian eigenvalues, (O) Otsu thresholding. White color means the output matches the fiber (True Positive), black means the voxel is correctly assigned as a polymer (True Negative), green means the algorithm assigned a voxel as a fiber while it was not (False Positive), orange means that a voxel belongs to a fiber, but the algorithm wrongly assigned it as a polymer (False Negative).

Results are more interesting for the LR data. Even the best possible threshold is far from being acceptable. As a result, the Otsu thresholding performs the worst. The Hessian based method has difficulties at low resolution, because fibers are getting closer to each other as the resolution decreases and therefore have more touching points. The trainable random forest works better than both Otsu and Hessian, but it is not able to find as good a representation as neural networks. Random forests are limited by features chosen and hand-designed beforehand. Neural networks are clearly outperforming the rest. See the results for both MR and LR in table 2.

**Table 2:** Performance of various methods on MR and LR data. The trainable methods were learned on Real MR 1, Synthetic MR 1 or Synthetic LR 1. Results in Dice score. The best threshold is known in advance from the groundtruth and is provided as the upper boundary for the Otsu thresholding. The highest achieved Dice scores for particular evaluation volumes are bolded.

| Evaluation volume | Training volume | Best threshold | Dice score of prediction | | | | |
|---|---|---|---|---|---|---|---|
| | | | Otsu | Hessian | RF | Deep 2D | Deep 3D |
| **Real MR 2** | Real MR 1 | 0.993 | 0.892 | 0.671 | 0.957 | ***0.986*** | 0.979 |
| | Synthetic MR 1 | | | | 0.796 | 0.903 | 0.881 |
| **Synthetic MR 2** | Synthetic MR 1 | 0.875 | 0.657 | 0.767 | 0.857 | 0.885 | ***0.898*** |
| **Synthetic LR 2** | Synthetic LR 1 | 0.505 | 0.289 | 0.374 | 0.536 | 0.750 | ***0.837*** |

## 6.2 Variations of the deep learning architecture

We compared four slightly different setups of deep networks. We trained the shallow and the deep version of the deep residual network architecture in both 2D and 3D variants (see figure 1) on real MR 1, synthetic MR 1 and synthetic LR 1 and evaluated on the other corresponding volumes. We discuss the performance and the capability of a network to generalize. The results are in table 3.

### 6.2.1 Highest score

To achieve the highest Dice score one has to train the network on an identical volume with similar statistics to the evaluation volume. For example, regardless the architecture, all networks achieved a Dice score of approx. 0.980 when trained on Real MR 1 and evaluated on Real MR 2. This is considerably better compared with standard methods. The same holds for training on Synthetic MR 1 and evaluating on Synthetic MR 2. When evaluating on the MR data, it seems that it is enough to use a 2D version of a network, with the 3D version not always working better. The deep version is usually only very slightly increasing the performance. Might be that the resolution is high enough that simple low level features already provide a high accuracy prediction. It is also easier to train it in a 2D setup, because of a lower number of trainable parameters. For LR data it is no longer the case. We find that already the shallow 2D version of the deep network is performing better than the other standard methods at LR. The deep 3D version is achieving the highest Dice score among the other architectures. This means, that accurately processing short glass fiber data at LR requires a richer feature representation.

### 6.2.2 Generalization

To evaluate the network capabilities to generalize we trained the network on volumes with different statistics than the evaluation volume. When trained on Synthetic MR 1 and evaluated on Real MR 1 or 2 the scores are still higher or similar to standard methods. The network trained on Synthetic MR 1 generalizes much better compared to training on Real MR 1 (and evaluated on Synthetic MR 1 or 2). That means when one does not know the statistics of the evaluation volume it is better to train the network on synthetic data than on real. Networks seem to overfit to certain directions when trained on Real MR 1, while fibers in our Synthetic MR data are uniformly oriented. When trained on Real MR 1 and evaluated on Synthetic MR 1 we can see that the 3D version is better at generalization. From this we conclude that the 3D features are more general, and harder to overfit. We did not note a gain in performance by using a deeper version of the 3D architecture.

Unfortunately, we do not have manually annotated data for real LR so we could not have done the same for real scans. Because of the same reason we cannot evaluate how well the network generalizes to real volumes (with different statistics).

**Table 3:** Comparison of different deep learning setups. Trained on real MR 1, synthetic MR 1 and LR 1. Results in Dice score. The highest achieved Dice scores for particular pair of evaluation and training volumes are bolded.

| Evaluation volume | Training volume | Dice score of prediction | | | |
|---|---|---|---|---|---|
| | | **Shallow 2D** | **Deep 2D** | **Shallow 3D** | **Deep 3D** |
| **Real MR 2** | Real MR 1 | 0.980 | ***0.986*** | 0.983 | 0.979 |
| | Synthetic MR 1 | ***0.928*** | 0.903 | 0.876 | 0.881 |
| **Synthetic MR 2** | Real MR 1 | 0.775 | 0.774 | ***0.801*** | 0.788 |
| | Synthetic MR 1 | 0.875 | 0.885 | 0.895 | ***0.898*** |
| **Real MR 1** | Synthetic MR 1 | ***0.929*** | 0.904 | 0.873 | 0.885 |
| **Synthetic MR 1** | Real MR 1 | 0.787 | 0.789 | ***0.818*** | 0.802 |
| **Synthetic LR 2** | Synthetic LR 1 | 0.630 | 0.750 | 0.803 | ***0.837*** |

### 6.2.3 Time evaluation

Lastly we take a look at the computational effort. We used single Nvidia GPU Titan X in Pascal architecture. Training and evaluation time for a 2D is considerably shorter than 3D. The same holds for deep vs shallow architecture. For 2D architecture, the network is predicting one slice at a time. In the 3D setup the network is predicting one patch at a time. These patches are overlapping and the final output is a mean output of patches. These timings could be optimized but it is out of scope of this paper. See table 4 for example timings for synthetic MR data.

**Table 4:** Time table of computational effort comparison of training and evaluation. Trained on a Synthetic MR 1 with 8,000 iterations and 3 patches per batch. It is evaluated on Synthetic MR 2 ($635 \times 603 \times 619$ [voxel]).

|  | **Shallow 2D** | **Deep 2D** | **Shallow 3D** | **Deep 3D** |
|---|---|---|---|---|
| **Training** | 173 [s] | 228 [s] | 400 [s] | 2,467 [s] |
| **Evaluation** | 17 [s] | 49 [s] | 184 [s] | 1,183 [s] |

## 3  Conclusions

We have shown that deep neural networks outperform other methods at LR and MR when properly trained and achieve the state-of-the-art. In the case of LR data, deep neural network is the only technique producing accurate results. In contrast to Otsu thresholding they are robust to changes in the histogram of intensity, and should produce results at least as good as Otsu. The benefit of using them in comparison with standard machine learning algorithms is the fact that they are capable of adapting the learnable features to the data. We also demonstrated that the network does not require a lot of real hand-annotated data in order to learn a working representation. It is possible to learn the network entirely on a synthetic volume. That means, if one does not want to spend time on annotating data, it is safe to use synthetic volumes at the cost of accuracy in segmentation and still have a higher accuracy than the standard methods.
The dataset and code can be found at *ipm-datasets.iwr.uni-heidelberg.de*